\documentclass[letterpaper]{article} 
\usepackage[draft]{aaai2026}
\usepackage{times}  
\usepackage{helvet}  
\usepackage{courier}  
\usepackage[hyphens]{url}  
\usepackage{graphicx} 
\usepackage{natbib}  
\usepackage{caption} 
\usepackage{algorithm}
\usepackage{algorithmic}

\usepackage{newfloat}
\usepackage{listings}
\DeclareCaptionStyle{ruled}{labelfont=normalfont,labelsep=colon,strut=off} 
\floatstyle{ruled}
\newfloat{listing}{tb}{lst}{}
\floatname{listing}{Listing}
\usepackage{amsfonts}
\usepackage{booktabs}
\usepackage{multirow}
\usepackage{comment}

\title{Exploitation Is All You Need... for Exploration}
\author {
    Micah Rentschler\textsuperscript{\rm 1},
    Jesse Roberts\textsuperscript{\rm 1}
}
\affiliations {
    \textsuperscript{\rm 1}Tennessee Technological University\\
    mrentschler@tntech.edu, jtroberts@tntech.edu
}

\begin{document}

\maketitle

\begin{abstract}
Ensuring sufficient exploration is a central challenge when training meta-reinforcement learning (meta-RL) agents to solve novel environments. Conventional solutions to the exploration–exploitation dilemma inject explicit incentives such as randomization, uncertainty bonuses, or intrinsic rewards to encourage exploration. In this work, we hypothesize that an agent trained solely to maximize a \emph{greedy} (exploitation-only) objective can nonetheless exhibit emergent exploratory behavior, provided three conditions are met: (1) \textbf{Recurring Environmental Structure}, where the environment features repeatable regularities that allow past experience to inform future choices; (2) \textbf{Agent Memory}, enabling the agent to retain and utilize historical interaction data; and (3) \textbf{Long-Horizon Credit Assignment}, where learning propagates returns over a time frame sufficient for the delayed benefits of exploration to inform current decisions. Through experiments in stochastic multi-armed bandits and temporally extended gridworlds, we observe that, when both structure and memory are present, a policy trained on a strictly greedy objective exhibits information-seeking exploratory behavior. We further demonstrate, through controlled ablations, that emergent exploration vanishes if either environmental structure or agent memory is absent (Conditions 1 \& 2). Surprisingly, removing long-horizon credit assignment (Condition 3) does not always prevent emergent exploration—a result we attribute to the \emph{pseudo-Thompson Sampling effect}. These findings suggest that, under the right prerequisites, exploration and exploitation need not be treated as orthogonal objectives but can emerge from a unified reward-maximization process.
\end{abstract}


\section{Introduction}

Consider children playing hide-and-go-seek repeatedly. The seeker, eager to win, pays attention to where friends have hidden in past rounds. She remembers which spots were successful, which were not, and adapts her searching accordingly. She explores less familiar places not out of intrinsic curiosity, but because the knowledge she gains may help her tag more hiders and score more points in the long run. Can machines do the same? Can they \textit{learn} to explore in order to exploit? That is the subject of this paper.

We define exploration as \textit{the selection of actions with the aim of acquiring information about the environment's dynamics or reward structure, thereby reducing epistemic uncertainty and enabling more effective decision-making.}

Exploration is ubiquitous in humans and animals, yet unlike drives such as hunger or pain, there is no clear, biologically plausible mechanism \underline{directly} incentivizing it. Neuroscience shows the brain balances exploration and exploitation through meta-learning, which relies on repeating tasks, memory, and rewards \cite{botvinick2018prefrontal}.

In reinforcement learning (RL), designers face a similar exploration–exploitation dilemma: how should an agent balance exploring unknown actions (potentially yielding future gain) and exploiting known rewards \cite{sutton2018reinforcement,thrun1992efficient}? Traditional RL methods address this by introducing explicit exploratory incentives such as $\epsilon$-greedy sampling \cite{sutton2018reinforcement}, Upper Confidence Bound (UCB) algorithms \cite{auer2002finite}, or intrinsic motivation mechanisms like curiosity-driven exploration \cite{pathak2019curiosity,burda2018large} that treat exploration as orthogonal to exploitation.

In contrast, we ask: \emph{Can exploration emerge organically from a purely greedy objective?} We hypothesize that it can—under three key conditions:

\begin{enumerate}
  \item \textbf{Recurring Environmental Structure.} The task repeats so early-acquired information remains valuable later.
  \item \textbf{Agent Memory.} The policy retains and utilizes past experiences across episodes.
  \item \textbf{Long-Horizon Credit Assignment.} Learning must connect information gathering to long-term payoffs.
\end{enumerate}

We empirically test this hypothesis via controlled ablation studies in both multi-armed bandits and temporally extended gridworlds. With all three conditions met, our results show that the greedy agent consistently exhibits information-seeking behavior even without explicit incentives. Removing structure or memory eliminates this effect. Remarkably in some contexts, emergent exploration persists even when training optimizes \underline{only} the immediate cumulative reward of the episode, suggesting a distinct alternate mechanism.

Our findings challenge the view that exploration and exploitation require separate objectives. In structured environments with memory, reward maximization alone enables effective exploration. This suggests shifting focus from exploration bonuses to memory-rich architectures that leverage repeated patterns, simplifying RL design and offering a biologically plausible explanation for exploration.

\section{Related Work}

The exploration—exploitation trade-off is a longstanding challenge in reinforcement learning (RL) and sequential decision making \cite{sutton2018reinforcement,lattimore2020bandit,thrun1992efficient}. Early approaches focused on randomization techniques, such as $\epsilon$-greedy or softmax action selection, as well as principled methods like Upper Confidence Bound (UCB) \cite{auer2002finite} and Thompson Sampling (TS) \cite{thompson1933likelihood}, which provide regret guarantees.

To extend tabular exploration to high-dimensional environments, density modeling and pseudo-count approaches were proposed \cite{bellemare2016unifying}. Hashing-based methods \cite{tang2017exploration} and successor representations \cite{machado2020countsr} have shown strong performance in complex domains by encouraging exploration through intrinsic bonuses.

Intrinsic motivation mechanisms, such as curiosity-driven exploration, reward agents for visiting novel or unpredictable states \cite{oudeyer2007intrinsic,pathak2019curiosity}. Random Network Distillation (RND) \cite{burda2019rnd} is a notable example, providing intrinsic bonuses in sparse-reward Atari games. Recent methods, such as SPIE \cite{yu2023spie}, combine counts and trajectory structure to target bottlenecks in exploration.

Meta-reinforcement learning (meta-RL) tackles rapid adaptation by learning across a distribution of related tasks \cite{duan2017rl2,zintgraf2020varibad}. Memory-rich architectures, such as recurrent neural networks \cite{hausknecht2015deep} and transformers \cite{melo2022transformers,chen2021decision,parisotto2020stabilizing}, allow policies to internalize exploratory strategies. In meta-RL, a recurrent policy successfully learns to explore via reward maximization alone \cite{duan2017rl2, wang2017learning}. VariBAD \cite{zintgraf2020varibad} applies Bayesian principles to guide implicit exploration, though with explicit modeling of uncertainty.

Neuroscientific evidence suggests that the brain’s meta-learning mechanisms manage the exploration–exploitation tradeoff by relying on specific prerequisites: recurring task structures, memory, and reward-driven learning processes \cite{botvinick2018prefrontal}. Notably, these prerequisites correspond closely to the conditions we have identified as essential for emergent exploration.

In addition to presenting new empirical findings, this study clarifies the boundaries and limitations of emergent exploration in meta-reinforcement learning. We systematically delineate and validate the specific conditions necessary for exploration to emerge from pure exploitation objectives, emphasizing the critical roles of recurring environmental structure and agent memory. Our analysis further raises new questions about long-term credit assignment, suggesting that its necessity may be context-dependent. To our knowledge, this is the first work to clearly articulate and formalize the preconditions for emergent exploration, demonstrating that it arises in multi-step environments when these preconditions are met. This provides a concrete foundation for ideas that have long been discussed in the field.

\section{Background}

\subsection{Repeated Markov Decision Processes (MDPs)}
To investigate when exploration emerges from pure exploitation, we consider a repeated-task setting: a fixed, partially observable Markov decision process (POMDP) instance
\[
M = (\mathcal{S}, \mathcal{A}, \mathcal{O}, P, \Omega, r, \gamma)
\]
is sampled from a distribution, and the agent interacts with this \emph{same} environment for multiple episodes. At the end of each episode, with probability $1/n$, a new parameterization of the environment is sampled, marking the start of a new task block. Thus, each block consists of a random number of episodes, geometrically distributed with mean of $n$, during which the environment's parameters remain fixed. Each episode begins from an initial state $s_0 \sim \rho$, while the transition dynamics $P$, observation kernel $\Omega$, and reward function $r$ are unchanged within a block. As a result, information acquired in early episodes—such as the location of a goal or the optimal arm—remains valuable in later episodes of the same block. This regime enables agents to accumulate and exploit knowledge across episodes.

\subsection{Meta-Reinforcement Learning (Meta-RL)}
Traditional RL trains agents to solve a single environment; meta-reinforcement learning (meta-RL) instead aims to create agents that rapidly adapt to new in-distribution tasks. In meta-RL, the agent is trained end-to-end—often via recurrent neural networks or transformers—to internalize optimization strategies within its policy, effectively ``learning to learn'' across episodes \cite{duan2017rl2,wang2017learning}. The agent’s memory (e.g., RNN hidden state or transformer context window) encodes past interactions, enabling the use of prior experience for fast adaptation within a block of repeated tasks.

Recently, meta-RL has more often employed transformers, which maintain memory by directly attending over the input sequence rather than through an explicit hidden state \cite{rentschler2025rltransformer,melo2022transformers}. The entire sequence of actions, observations, and rewards is tokenized and fed to the transformer as input. Causal transformers restrict attention to past tokens, supporting autoregressive sequence modeling. Pretraining and subsequent finetuning can enhance adaptability and context sensitivity.

\section{Methodology}
\label{sec:methodology}

Our experiments are designed to empirically test the hypothesis that exploration can emerge from pure exploitation objectives given three jointly necessary factors are met: recurring environmental structure, agent memory, and a sufficiently long temporal horizon. We perform controlled ablation studies, systematically removing each factor to observe its impact on agent behavior.

\subsection{Environments}
We evaluate our hypothesis using both multi-armed bandit problems and multi-step gridworlds:

\begin{itemize}
    \item \textbf{Multi-Armed Bandits:} In the $K$-armed bandit setting, each arm provides rewards from a stationary distribution (in our experiments a Bernoulli distribution). The agent selects arms across multiple episodes, with reward parameters fixed within each task block. At the end of each episode, with probability $1/n$, a new block begins and new reward distributions are sampled; otherwise, the current block continues. Thus, the block length is a geometric random variable with mean $n$. By varying $n$, we control the degree of recurring structure: small $n$ yields little structure, while larger $n$ allows information acquired early to generalize to later episodes.
    \item \textbf{Gridworlds:} For longer-horizon tasks, we use variants of the Frozen Lake gridworld. Agents must navigate from a start state to a goal, receiving reward only upon success. We modified the typical reward for success to $1/t$ where $t$ is the number of time steps taken to reach the goal, giving the agent more reward the faster it reaches the goal. The environment remains fixed for a randomly determined number of episodes per block, with each episode terminating the block with probability $1/n$ (i.e., mean block length $n$), after which a new grid is generated. This setup allows us to test whether agents are willing to forgo short-term rewards to gather information that aids later performance.
\end{itemize}

\begin{figure}[ht]
    \centering
    \includegraphics[width=1\linewidth]{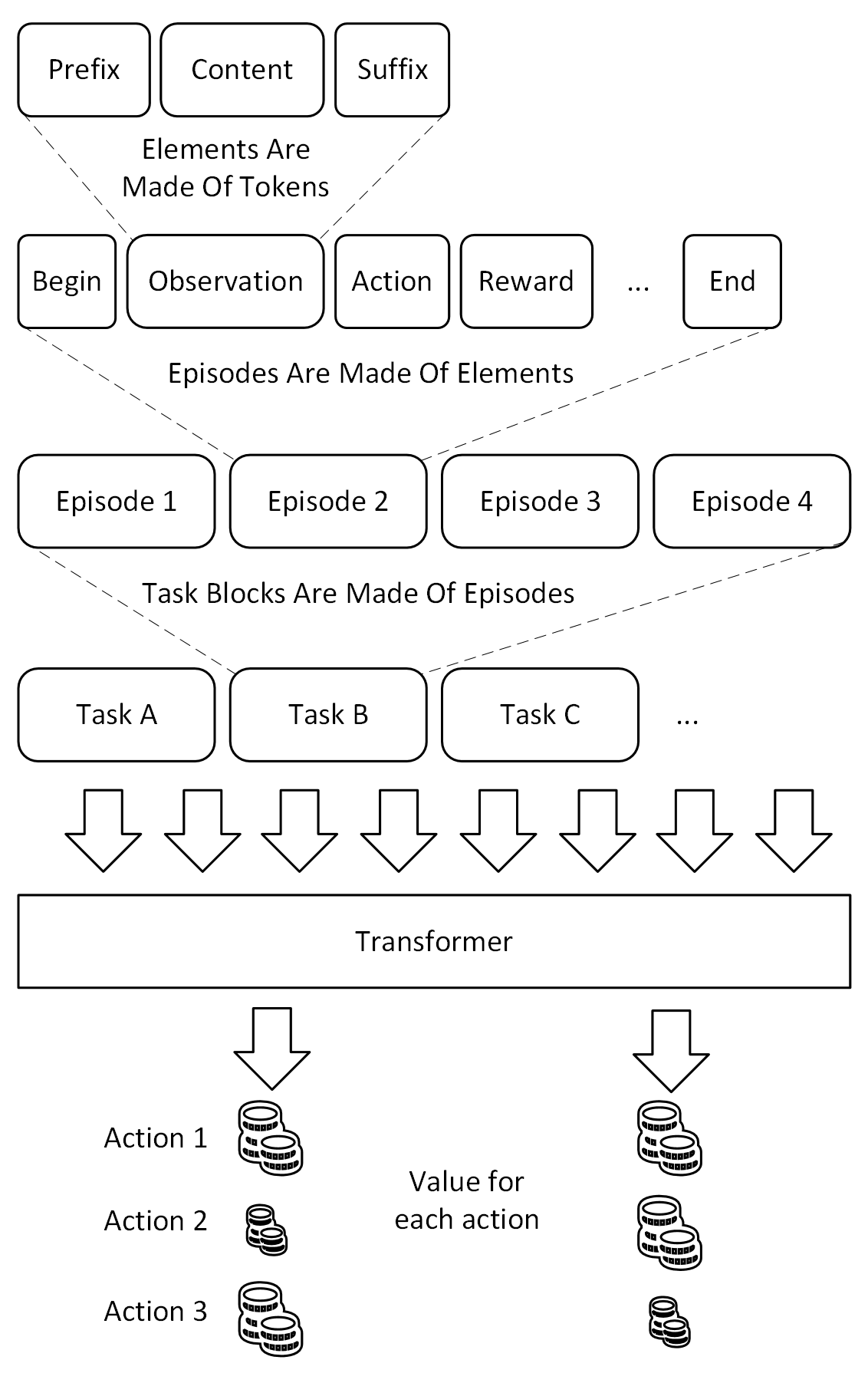}
    \caption{Setup for DQN meta-RL learning with a transformer. Tokens make up elements, which comprise episodes, which in turn make up task blocks. This is fed to the transformer which is trained to output the value of each action.}
    \label{fig:transformer_setup}
\end{figure}

\subsection{Agent Architecture}
Agents are implemented using transformer-based value functions, following \citet{rentschler2025rltransformer}. The model processes the most recent $X$ tokens (composed of actions, observations, and rewards), which determines memory capacity. Smaller $X$ limits memory to recent events, while larger $X$ enables cross-episode retention. We use the pretrained Llama 3.2 3B model with LoRA adapters ($r_{lora}=32$, $\alpha_{lora}=32$) for efficient fine-tuning.

\subsection{Training Procedure}

Agents are trained with the Deep Q-Network (DQN) algorithm \cite{mnih2015human}, using a step discount factor $\gamma_{step}$ to set the effective temporal credit assignment. At the end of each episode, we use the episode discount factor $\gamma_{episode}$. The value function is reset at block boundaries, so the objective is to maximize total expected discounted reward over each block. A delayed target network is necessary to stabilize learning. We use Polyak averaging to update the weights of the the target network with a factor $\alpha=0.1$.

We trained on a single H200 GPU. Training is performed for 400 iterations in bandit environments and 1200 iterations in gridworlds unless otherwise noted. Training is conducted offline: $M$ data streams are generated by running environments with uniformly sampled actions until each stream contains at least $20X$ tokens. In each training iteration, a fixed number of streams (32 for $X=1024$, 128 for $X=256$, 256 for $X=128$, 512 for $X=64$) are sampled to make a batch, thus keeping the number of tokens per batch constant for different context windows. Action masking is applied to ensure only valid actions are included in the DQN loss.

\subsection{Testing Procedure}

Testing is performed on either a H200 or RTX3090 GPU. Before testing, we fill the context of the transformer with experience trajectories generated by taking random actions (i.e. seeding the context). Then, we let the transformer take actions for a predefined number of episodes.

\subsection{Ablation Studies}
To isolate the contribution of each precondition, we systematically vary:

\begin{itemize}
    \item \textbf{Recurring Structure:} Controlled by the mean block length $n$ (large $n$ is equivalent to a high degree of recurrence, while small $n$ indicates limited repetitiveness).
    \item \textbf{Agent Memory:} Controlled by transformer context window $X$ (small $X$ limits memory, large $X$ enables cross-episode memory).
    \item \textbf{Temporal Horizon:} Controlled by discount factor $\gamma_{episode}$ (lower values shorten the planning horizon; higher values enable long-term credit assignment).
\end{itemize}

Unless otherwise specified, all other hyperparameters and architectural details are held constant.

Performance is assessed by the extent to which agents take actions that forgo immediate reward in favor of gathering information that benefits future episodes in the same block. This exploratory behavior is best measured by the total average reward over the whole episode. Agents that explore are more likely to find better strategies and achieve higher total rewards while agents that do not explore will tend to exploit randomly encountered strategies, leading to a low expected reward. 

\subsection{Baselines}
Our goal is not to outperform standard baselines, but to demonstrate that competitive performance can be achieved without explicit exploration mechanisms.

For bandit environments, we compare to Thompson sampling, $\epsilon$-greedy, oracle, and random strategies.

In the gridworld setting, where standard baselines are less established, we evaluate agent performance using both average reward and state visitation counts as indicators of exploration. For reference, we compare these metrics to those achieved by an oracle policy and a random policy, providing context for interpreting the effectiveness of our approach.

\section{Experiments}

We evaluate our hypothesis—that exploration can emerge from pure exploitation given the right structural conditions—using controlled ablation studies in multi-armed bandit and gridworld tasks. Across all experiments, we systematically vary: the degree of environment recurrence ($n$), agent memory capacity ($X$), and the temporal credit assignment horizon ($\gamma_{episode}$).

\subsection{Experimental Setup}

For both bandit and gridworld experiments, we follow the repeated-task setup described in Section~\ref{sec:methodology}. 

All results are averaged over multiple seeds: 10,000 runs for baselines and 1,000 for meta-RL in the bandit task, and 1,000 runs for baselines and 100 for meta-RL in gridworld. We also report 95\% confidence intervals.

To facilitate comparison and interpretation, we linearly normalize all reported rewards: a value of 1 denotes performance of an oracle agent, while 0 indicates the average performance of a random agent under identical conditions.

\subsection{Multi-Armed Bandit Results}

\begin{figure}[t]
    \centering
    \includegraphics[width=1\linewidth]{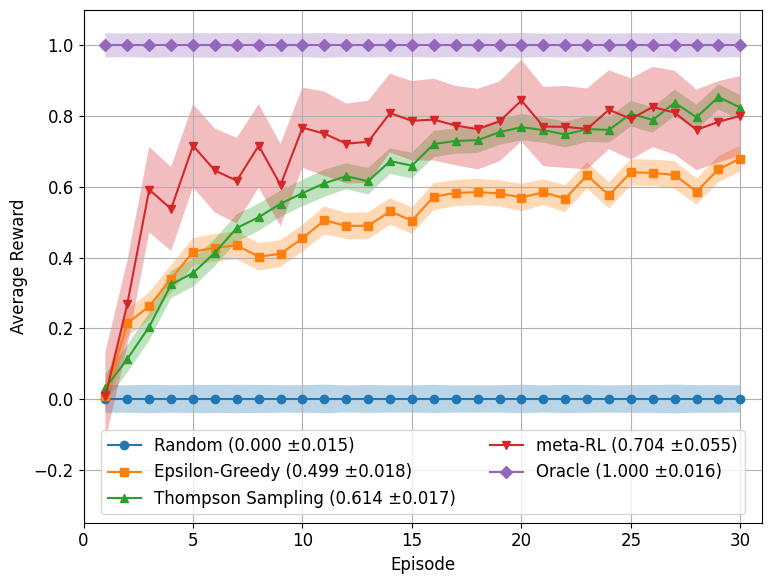}
    \caption{
        Reward per episode in the 3-arm bandit environment, comparing meta-RL ($n=30$ $X=1024$ $\gamma_{episode}=0.9$) to baseline strategies. Shaded areas denote 95\% confidence intervals. The meta-RL agent exceeds Thompson Sampling and $\epsilon$-greedy baselines, demonstrating its ability to explore in the absence of explicit incentives.
    }
    \label{fig:reward_bandit}
\end{figure}

Figure~\ref{fig:reward_bandit} shows the performance of the meta-RL agent in a 3-arm bandit environment. The agent’s returns exceed those of the Thompson Sampling and $\epsilon$-greedy baselines. This result is expected, as both Thompson Sampling and $\epsilon$-greedy are designed to maximize asymptotic reward; in finite-horizon settings, there is room for other approaches to achieve better performance \cite{duan2017rl2}.

\paragraph{Effect of Recurring Structure}

Table~\ref{tab:var_episode_length} shows that higher mean block lengths ($n$)—i.e., greater environment recurrence—enable agents to leverage early exploration for improved performance in later episodes. Performance drops sharply for smaller $n$, confirming that recurring structure is necessary for emergent exploration.

\begin{table}[H]
    \centering
    \begin{tabular}{lccc}
        \toprule
        $n$ & $\epsilon$-Greedy & TS & Meta-RL \\
        \midrule
        30 & 0.499 ±0.018 & 0.614 ±0.017 & 0.704 ±0.055 \\
        10 & 0.354 ±0.020 & 0.339 ±0.019 & 0.509 ±0.064 \\
        3 & 0.147 ±0.027 & 0.092 ±0.025 & 0.325 ±0.082 \\
        1 & -0.083 ±0.041 & -0.066 ±0.041 & 0.043 ±0.130 \\
        \bottomrule
    \end{tabular}
    \caption{
        Cumulative reward per task block as a function of mean block length $n$ in the 3-armed bandit environment.
    }
    \label{tab:var_episode_length}
\end{table}

\paragraph{Effect of Memory Capacity}

Table~\ref{tab:var_context_length} demonstrates that reducing the agent's memory capacity (context window $X$) leads to a sharp decline in cumulative reward. Below a critical threshold, exploration fails to emerge, confirming that sufficient memory is essential.

\begin{table}[H]
    \centering
    \begin{tabular}{lccc}
        \toprule
        $X$ & $\epsilon$-Greedy & TS & Meta-RL \\
        \midrule
        1024
        & \multirow{5}{*}{0.499 ±0.018} 
        & \multirow{5}{*}{0.614 ±0.017} 
        & 0.704 ±0.055 \\
        256 & & & 0.792 ±0.054 \\
        128 & & & 0.574 ±0.062 \\
        64 & & & 0.547 ±0.060 \\
        32 & & & -0.052 ±0.099 \\
        \bottomrule
    \end{tabular}
    \caption{
        Cumulative reward per task block as a function of transformer context window $X$ in the 3-armed bandit.
    }
    \label{tab:var_context_length}
\end{table}

\paragraph{Effect of Temporal Credit Assignment}

\begin{figure}[t]
    \centering
    \includegraphics[width=1.0\linewidth]{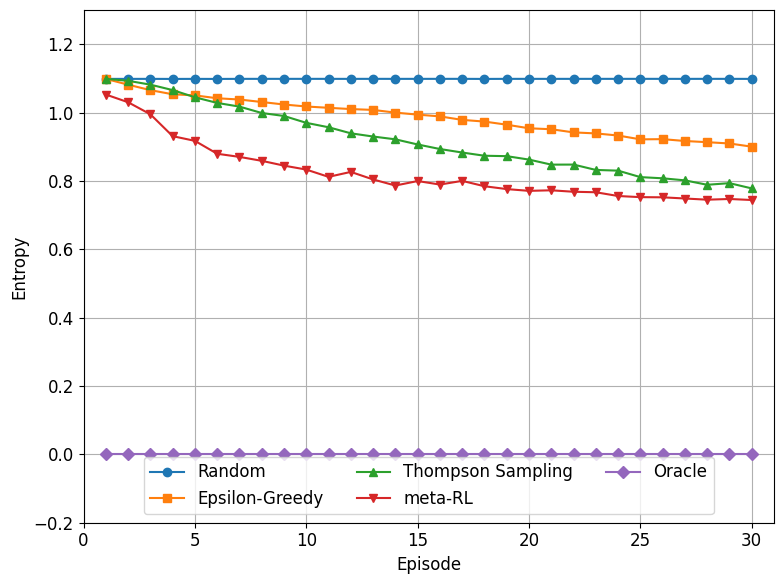}
    \caption{
        Entropy of the action distribution derived from multiple trials with identical environment parameters, seeded with randomly generated episodes. The meta-RL agent's (trained with $\gamma_{episode}=0$) actions are initially stochastic (exploratory), becoming more deterministic (exploitative)  as it accumulates experience, demonstrating that the agent can output pseudo-stochastic samples conditioned on contexts.
    }
    \label{fig:entropy_bandit}
\end{figure}

When the temporal horizon is ablated ($\gamma_{episode}=0$), surprisingly, meta-RL agents continue to display exploratory behavior, without appreciable reduction in performance. Figure~\ref{fig:entropy_bandit} shows the entropy of the action distribution, measured by running multiple trials with identical environment parameters while seeding the context with randomly generated episodes. The agent starts with random actions and becomes more deterministic as it accumulates experience. This shows that the agent can output pseudo-stochastic samples conditioned on long, complex, and chaotic contexts.

\subsection{Gridworld Results}

\begin{figure}[t]
    \centering
    \includegraphics[width=1\linewidth]{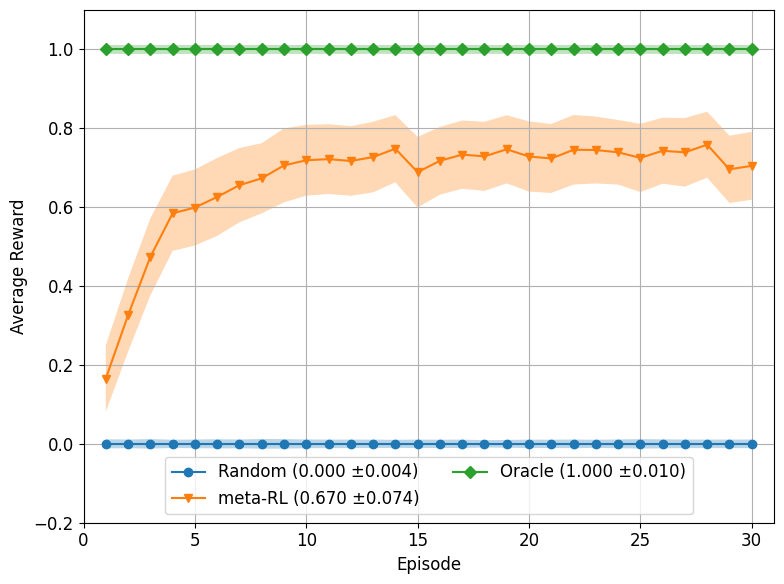}
    \caption{
        Reward per episode in the Frozen Lake gridworld, comparing meta-RL ($n=30$ $X=1024$ $\gamma_{episode}=0.9$) to random and oracle strategies. Shaded areas denote 95\% confidence intervals. The meta-RL agent achieves significant gains, demonstrating that it gathers information (exploration) that is later used to gather rewards (exploitation).
    }
    \label{fig:reward_gridworld}
\end{figure}

Figure~\ref{fig:reward_gridworld} shows the performance of the meta-RL agent in the Frozen Lake gridworld. The agent achieves decent performance, closing the gap between a random policy and the oracle (always-win policy) by 55.2\%.

\paragraph{Effect of Recurring Structure}

As with bandits, increasing block length $n$ in gridworlds yields higher cumulative reward (Table~\ref{tab:var_episode_length_gridworld}). The agent exploits repeated structure to improve over time.

\begin{table}[H]
    \centering
    \begin{tabular}{lc}
        \toprule
        $n$ & Meta-RL \\
        \midrule
        30 & 0.670 ±0.074 \\
        10 & 0.441 ±0.070 \\
        3  & 0.254 ±0.071 \\
        1  & -0.050 ±0.004 \\
        \bottomrule
    \end{tabular}
    \caption{
        Cumulative reward per task block in Frozen Lake gridworld as a function of mean block length $n$.
    }
    \label{tab:var_episode_length_gridworld}
\end{table}

\paragraph{Effect of Memory Capacity}

Reducing the context window $X$ also decreases agent performance in gridworlds, though the critical threshold for collapse occurs at higher $X$ than in bandits due to the increased episode length (Table~\ref{tab:var_context_length_gridworld}).

\begin{table}[H]
    \centering
    \begin{tabular}{lc}
        \toprule
        $X$ & Meta-RL \\
        \midrule
        1024 & 0.670 ±0.074 \\
        256  & 0.120 ±0.056 \\
        128  & 0.047 ±0.026 \\
        64   & -0.001 ±0.033 \\
        \bottomrule
    \end{tabular}
    \caption{
        Cumulative reward per task block as a function of memory window $X$ in gridworld.
    }
    \label{tab:var_context_length_gridworld}
\end{table}

\paragraph{Effect of Temporal Credit Assignment}

Our results show that while meta-RL agents can exhibit exploratory behavior in bandit tasks even with $\gamma_{episode} = 0$, in gridworld environments, increasing the discount factor from zero to a nonzero value leads to a moderate improvement in overall performance (which is indicative of emergent exploration), with the average reward rising from 0.408 ±0.089 to 0.670 ±0.074.

\paragraph{State Visitation}

\begin{figure}[bp]
    \centering
    \begin{tabular}{@{} c l @{}}
        \includegraphics[height=0.28\textheight]{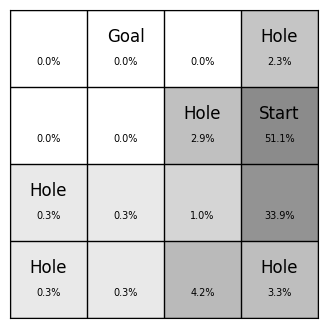} &
        \raisebox{0.145\textheight}{Early} \\
        \includegraphics[height=0.28\textheight]{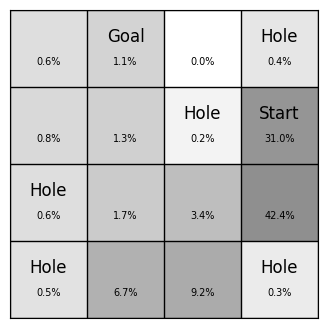} &
        \raisebox{0.145\textheight}{Middle} \\
        \includegraphics[height=0.28\textheight]{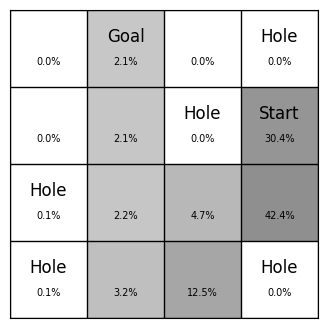} &
        \raisebox{0.145\textheight}{Late} \\
    \end{tabular}
    \caption{
        State visitation heatmaps in the same gridworld (top: episodes 1--3; middle: 4--7; bottom: 8--30; over ten trials). Early episodes show wide exploration; later episodes focus on efficient exploitation of discovered paths to goals. This progression illustrates emergent exploration followed by exploitation.
    }
    \label{fig:state_visitation}
\end{figure}

Figure~\ref{fig:state_visitation} shows the state visitation distribution across block progression aggregated over 10 trials in the same environment with different randomly seeded contexts (percentages represent the relative number of time steps spent in a particular state). Early on, agents visit a broad range of states. As training progresses, visitation becomes concentrated along paths leading to the nearest goal, and agents appear to avoid hazardous states such as holes. Notably, exploration seems to expand outward from the starting position, gradually covering more distant regions.

It is important to note that not all trials follow this typical pattern. In some instances, agents fail to find the goal entirely, while in others, they discover the goal early by chance and subsequently cease exploring further. The most significant failure cases occur when agents waste time executing sequences of actions that result in running into walls (evidenced by high state occupation). Nevertheless, the overall trends observed are consistent with incremental progress toward effective exploration.

\section{Discussion}
\label{sec:discussion}

Our empirical results provide compelling evidence that, contrary to the standard dichotomy in reinforcement learning literature, exploration can arise as an emergent consequence of pure exploitation. We show that, when the environment exhibits recurring structure and the agent possesses adequate memory, even a purely greedy policy reliably produces information-seeking, exploratory behavior. This is demonstrated empirically across both bandit and temporally extended gridworld tasks.

\subsection{Agent Memory and Environment Repetition}

A key insight from our ablation studies is the necessity of the first two preconditions - recurring environmental structure and agent memory - for emergent exploration. When either is removed, exploratory behavior collapses: the agent remains myopic, failing to gather information that could improve future returns. This directly supports our hypothesis that both recurrent structure and memory are necessary ingredients for the emergence of organic exploration.

\subsection{Long-Term Credit Assignment}

Perhaps most surprisingly, our experiments with multi-armed bandits reveal that long-term credit assignment is \textbf{not essential} for exploration to emerge, in seeming contradiction to our initial hypothesis. In bandit domains, even when the temporal horizon is removed ($\gamma_{episode}=0$), the agent continues to exhibit behavior similar to state-of-the-art baselines. This suggests that, under some circumstances, greedy policies with memory can approximate efficient exploration strategies without explicitly propagating delayed reward.

To understand this, recall that with $\gamma_{episode}=0$, the action-value function in DQN approximates the immediate reward received for a chosen action. If the value function could generate a sample from the \emph{distribution} of rewards likely to be received, then the DQN algorithm would be doing Thompson Sampling (sample the distribution, take the most promising action, update the belief). However, standard neural networks generally approximate the mean reward, not the full distribution, seemingly making this equivalence impossible.

However, recent studies suggest that in-context learning with transformers can produce outputs that effectively sample from a pseudo-distribution, even exhibiting pseudo-random behavior conditioned on context \cite{hataya2024transformers}. We therefore propose that the mechanism by which exploration emerges in our setting when $\gamma_{episode}=0$ is not due to long-term credit assignment, but rather is an artifact of the transformer's ability to generate context-dependent samples from an approximate reward distribution. When combined with the DQN update rule, this can closely approximate Thompson Sampling, even without explicit Bayesian modeling. We term this phenomenon \textbf{pseudo-Thompson Sampling} (pseudo-TS).

There is other evidence that leads us to believe in the possibility of this mechanism. Bayesian dynamic programming \cite{strens2000bayesian}, a model-based algorithm, demonstrated an ability to explore the environment without any auxiliary reward by sampling a model of the environment from a distribution and learning the optimal policy for that model with dynamic programming. This is, in effect, Thompson Sampling for model-based RL \cite{tziortziotis2013linear}. \citet{osband2019deep} showed in deep RL that one can induce efficient exploration without any added reward bonus by using randomized value functions, which is conceptually what we are claiming is happening, although the source of randomization is different.

Moreover, this pseudo-Thompson Sampling (pseudo-TS) effect is not necessarily limited to stateless environments. If the action-value function can accurately represent the distributional value of actions far into the future, then Thompson-style sampling from such a value function could, in principle, induce exploratory actions even in temporally extended tasks.

However, our gridworld experiments show that a non-zero discount factor is indeed beneficial for emergent exploration and improved cumulative reward. This suggests that the effectiveness of pseudo-TS is closely tied to how well the value function captures the distribution of future rewards. As the temporal reasoning horizon grows and estimating the value function becomes more challenging, pseudo-TS becomes less accurate and cannot fully substitute for a learning signal that directly motivates information-seeking actions with long-term payoff. Thus, in multi-step environments, an appropriately chosen discount factor remains important for supporting effective exploration.

\subsection{Overall Takeaway}

From a practical perspective, our results suggest that RL algorithm designers may \textit{benefit more from focusing on generalizing with meta-RL conditioned on task history}, rather \textit{than on increasingly sophisticated exploration bonuses}. In environments with significant recurring structure, this approach can naturally produce effective exploration, simplifying both the design and tuning of RL algorithms. The bitter lesson that general methods leveraging massive computation are ultimately the most effective seems to have struck again \cite{sutton2019bitter}.

\subsection{Limitations}

Nonetheless, there are important limitations. Our study is focused on environments with explicit recurring structure and assumes sufficient agent memory and context length. While we claim these conditions are necessary, they may not be sufficient; exploration may fail to arise for a variety of other reasons. Additionally, the mechanism underlying pseudo-Thompson Sampling is empirical and not theoretically established, and our results may not generalize to more complex or high-dimensional tasks. Further research is needed to assess the robustness and scalability of emergent exploration in broader domains.

\section{Conclusion}

We have shown that exploration can emerge naturally from pure exploitation objectives, provided two core conditions are met: (1) the environment contains recurring, exploitable structure, and (2) the agent has sufficient memory to retain and utilize past experience. Contrary to conventional wisdom, our experiments suggest that explicit exploration bonuses or intrinsic rewards are not always required; instead, information-seeking behavior can arise naturally without external incentives.

While we initially hypothesized that a third condition—long-term temporal credit assignment—would be necessary for emergent exploration, our ablation studies reveal a more nuanced picture. In some cases, particularly when the value function can effectively represent the distribution over possible returns, exploration can arise even without explicit long-horizon credit assignment. This phenomenon is most plausibly explained by a pseudo-Thompson Sampling effect, enabled by transformer-based architectures that can condition on diverse contexts and approximate distributional value functions. In essence, if a generative model can produce samples from the value function's distribution, exploration can emerge “for free.” However, when the model's distributional estimates are insufficient—such as in more complex, temporally extended tasks—a nonzero discount factor and longer temporal horizon become essential for supporting effective exploration.

These findings suggest a potential paradigm shift: rather than treating exploration and exploitation as fundamentally orthogonal, we can view exploration as an emergent property of reward-maximizing behavior in structured, memory-rich contexts. This emphasizes the need for meta-reinforcement learning architectures that support persistent memory in order to leverage environmental regularities.

We hope these results encourage continued investigation into how context, memory, long horizons, and distributional modeling contribute to emergent exploration, ultimately bringing us closer to agents that explore simply because it is the best way to maximize reward.

\appendix

\bibliography{aaai2026}

\begin{thebibliography}{28}
\providecommand{\natexlab}[1]{#1}

\bibitem[{Auer, Cesa-Bianchi, and Fischer(2002)}]{auer2002finite}
Auer, P.; Cesa-Bianchi, N.; and Fischer, P. 2002.
\newblock Finite-time analysis of the multiarmed bandit problem.
\newblock \emph{Machine Learning}, 47(2-3): 235--256.

\bibitem[{Bellemare et~al.(2016)Bellemare, Srinivasan, Ostrovski, Schaul, Saxton, and Munos}]{bellemare2016unifying}
Bellemare, M.~G.; Srinivasan, S.; Ostrovski, G.; Schaul, T.; Saxton, D.; and Munos, R. 2016.
\newblock Unifying count-based exploration and intrinsic motivation.
\newblock In \emph{Advances in Neural Information Processing Systems 29 (NeurIPS)}, 1471--1479.

\bibitem[{Botvinick et~al.(2018)Botvinick, Wang, Kwisthout, and et~al.}]{botvinick2018prefrontal}
Botvinick, M.; Wang, J.~X.; Kwisthout, J.; and et~al. 2018.
\newblock Prefrontal cortex as a meta-reinforcement learning system.
\newblock \emph{Nature Neuroscience}, 21(6): 860--868.

\bibitem[{Burda et~al.(2018)Burda, Edwards, Pathak, Storkey, Darrell, and Efros}]{burda2018large}
Burda, Y.; Edwards, H.; Pathak, D.; Storkey, A.; Darrell, T.; and Efros, A.~A. 2018.
\newblock Large-scale study of curiosity-driven learning.
\newblock \emph{arXiv preprint arXiv:1808.04355}.

\bibitem[{Burda et~al.(2019)Burda, Edwards, Pathak, Storkey, Darrell, and Efros}]{burda2019rnd}
Burda, Y.; Edwards, H.; Pathak, D.; Storkey, A.; Darrell, T.; and Efros, A.~A. 2019.
\newblock Exploration by random network distillation.
\newblock In \emph{International Conference on Learning Representations (ICLR)}.

\bibitem[{Chen et~al.(2021)Chen, Lu, R.~Kumar, Zhou, Bavarian et~al.}]{chen2021decision}
Chen, L.; Lu, K.; R.~Kumar, A.; Zhou, H.; Bavarian, M.; et~al. 2021.
\newblock Decision Transformer: Reinforcement Learning via Sequence Modeling.
\newblock In \emph{Advances in Neural Information Processing Systems (NeurIPS)}.

\bibitem[{Duan et~al.(2017)Duan, Schulman, Chen, Bartlett, Sutskever, and Abbeel}]{duan2017rl2}
Duan, Y.; Schulman, J.; Chen, X.; Bartlett, P.~L.; Sutskever, I.; and Abbeel, P. 2017.
\newblock {RL}$^2$: Fast reinforcement learning via slow reinforcement learning.
\newblock In \emph{International Conference on Learning Representations (ICLR)}.

\bibitem[{Hataya and Imaizumi(2024)}]{hataya2024transformers}
Hataya, R.; and Imaizumi, M. 2024.
\newblock Transformers as Stochastic Optimizers.
\newblock In \emph{ICML 2024 Workshop on In-Context Learning}.

\bibitem[{Hausknecht and Stone(2015)}]{hausknecht2015deep}
Hausknecht, M.; and Stone, P. 2015.
\newblock Deep Recurrent Q-Learning for Partially Observable MDPs.
\newblock In \emph{AAAI Conference on Artificial Intelligence}.

\bibitem[{Lattimore and Szepesv{\'a}ri(2020)}]{lattimore2020bandit}
Lattimore, T.; and Szepesv{\'a}ri, C. 2020.
\newblock \emph{Bandit Algorithms}.
\newblock Cambridge University Press.

\bibitem[{Machado, Bellemare, and Bowling(2020)}]{machado2020countsr}
Machado, M.~C.; Bellemare, M.~G.; and Bowling, M. 2020.
\newblock Count-Based Exploration with the Successor Representation.
\newblock \emph{Proceedings of the AAAI Conference on Artificial Intelligence}, 34(04): 5125--5133.

\bibitem[{Melo(2022)}]{melo2022transformers}
Melo, L.~C. 2022.
\newblock Transformers are meta-reinforcement learners.
\newblock In \emph{International Conference on Machine Learning (ICML)}, 15340--15359.

\bibitem[{Mnih et~al.(2015)Mnih, Kavukcuoglu, Silver, Rusu, Veness, Bellemare, Graves, Riedmiller, Fidjeland, Ostrovski et~al.}]{mnih2015human}
Mnih, V.; Kavukcuoglu, K.; Silver, D.; Rusu, A.~A.; Veness, J.; Bellemare, M.~G.; Graves, A.; Riedmiller, M.; Fidjeland, A.~K.; Ostrovski, G.; et~al. 2015.
\newblock Human-level control through deep reinforcement learning.
\newblock \emph{Nature}, 518(7540): 529--533.

\bibitem[{Osband et~al.(2019)Osband, Van~Roy, Russo, and Wen}]{osband2019deep}
Osband, I.; Van~Roy, B.; Russo, D.~J.; and Wen, Z. 2019.
\newblock Deep exploration via randomized value functions.
\newblock \emph{Journal of Machine Learning Research}, 20(124): 1--62.

\bibitem[{Oudeyer and Kaplan(2007)}]{oudeyer2007intrinsic}
Oudeyer, P.-Y.; and Kaplan, F. 2007.
\newblock Intrinsic motivation systems for autonomous mental development.
\newblock \emph{IEEE Transactions on Evolutionary Computation}, 11(2): 265--286.

\bibitem[{Parisotto et~al.(2020)Parisotto, Song, Rae, Pascanu, Guez, and et~al.}]{parisotto2020stabilizing}
Parisotto, E.; Song, F.; Rae, J.~W.; Pascanu, R.; Guez, A.; and et~al. 2020.
\newblock Stabilizing Transformers for Reinforcement Learning.
\newblock In \emph{International Conference on Machine Learning (ICML)}.

\bibitem[{Pathak et~al.(2019)Pathak, Agrawal, Efros, Darrell, and Malik}]{pathak2019curiosity}
Pathak, D.; Agrawal, P.; Efros, A.~A.; Darrell, T.; and Malik, J. 2019.
\newblock Curiosity-driven exploration by self-supervised prediction.
\newblock In \emph{International Conference on Learning Representations (ICLR)}.

\bibitem[{Rentschler and Roberts(2025)}]{rentschler2025rltransformer}
Rentschler, M.; and Roberts, J. 2025.
\newblock {RL} + Transformer = A General-Purpose Problem Solver.
\newblock In Kamalloo, E.; Gontier, N.; Lu, X.~H.; Dziri, N.; Murty, S.; and Lacoste, A., eds., \emph{Proceedings of the 1st Workshop for Research on Agent Language Models (REALM 2025)}, 401--410. Vienna, Austria: Association for Computational Linguistics.
\newblock ISBN 979-8-89176-264-0.

\bibitem[{Strens(2000)}]{strens2000bayesian}
Strens, M. 2000.
\newblock A Bayesian framework for reinforcement learning.
\newblock In \emph{International Conference on Machine Learning (ICML)}, 943--950.

\bibitem[{Sutton(2019)}]{sutton2019bitter}
Sutton, R. 2019.
\newblock The bitter lesson.
\newblock \emph{Incomplete Ideas (blog)}, 13(1): 38.

\bibitem[{Sutton and Barto(2018)}]{sutton2018reinforcement}
Sutton, R.~S.; and Barto, A.~G. 2018.
\newblock \emph{Reinforcement Learning: An Introduction}.
\newblock MIT Press, 2nd edition.

\bibitem[{Tang et~al.(2017)Tang, Houthooft, Foote, Stooke, Chen, Duan, Schulman, De~Turck, and Abbeel}]{tang2017exploration}
Tang, H.; Houthooft, R.; Foote, D.; Stooke, A.; Chen, X.; Duan, Y.; Schulman, J.; De~Turck, F.; and Abbeel, P. 2017.
\newblock \#Exploration: A Study of Count-Based Exploration for Deep Reinforcement Learning.
\newblock In \emph{Advances in Neural Information Processing Systems 30 (NeurIPS)}, 2750--2759.

\bibitem[{Thompson(1933)}]{thompson1933likelihood}
Thompson, W.~R. 1933.
\newblock On the likelihood that one unknown probability exceeds another in view of the evidence of two samples.
\newblock \emph{Biometrika}, 25(3/4): 285--294.

\bibitem[{Thrun(1992)}]{thrun1992efficient}
Thrun, S. 1992.
\newblock Efficient exploration in reinforcement learning.
\newblock In \emph{Advances in Neural Information Processing Systems (NeurIPS)}, 433--440.

\bibitem[{Tziortziotis, Dimitrakakis, and Blekas(2013)}]{tziortziotis2013linear}
Tziortziotis, N.; Dimitrakakis, C.; and Blekas, K. 2013.
\newblock Linear Bayesian Reinforcement Learning.
\newblock In \emph{International Joint Conference on Artificial Intelligence (IJCAI)}, 1721--1728.

\bibitem[{Wang et~al.(2017)Wang, Kurth-Nelson, Tirumala, Soyer, Leibo, Munos, Blundell, Kumaran, and Botvinick}]{wang2017learning}
Wang, J.~X.; Kurth-Nelson, Z.; Tirumala, D.; Soyer, H.; Leibo, J.~Z.; Munos, R.; Blundell, C.; Kumaran, D.; and Botvinick, M. 2017.
\newblock Learning to reinforcement learn.
\newblock In \emph{Proceedings of the 39th Annual Conference of the Cognitive Science Society (CogSci)}.

\bibitem[{Yu et~al.(2023)Yu, Burgess, Sahani, and Gershman}]{yu2023spie}
Yu, C.; Burgess, N.; Sahani, M.; and Gershman, S.~J. 2023.
\newblock Successor-Predecessor Intrinsic Exploration.
\newblock In \emph{Advances in Neural Information Processing Systems (NeurIPS)}.

\bibitem[{Zintgraf et~al.(2020)Zintgraf, Shiarlis, Igl, Schulze, Gal, Hofmann, and Whiteson}]{zintgraf2020varibad}
Zintgraf, L.; Shiarlis, K.; Igl, M.; Schulze, S.; Gal, Y.; Hofmann, K.; and Whiteson, S. 2020.
\newblock {VariBAD: A Very Good Method for Bayes-Adaptive Deep RL via Meta-Learning}.
\newblock In \emph{International Conference on Learning Representations (ICLR)}.

\end{thebibliography}

\end{document}